\documentclass[lettersize,journal]{IEEEtran}
\usepackage{amsmath,amsfonts}
\usepackage{algorithmic}
\usepackage{algorithm}
\usepackage{array}
\usepackage[caption=false,font=normalsize,labelfont=sf,textfont=sf]{subfig}
\usepackage{textcomp}
\usepackage{stfloats}
\usepackage{url}
\usepackage{verbatim}
\usepackage{graphicx}
\usepackage{cite}

\usepackage{booktabs}
\usepackage{multirow}

\usepackage{amsmath}
\usepackage{color}
 
\usepackage{amssymb}
\usepackage{pifont} 
\graphicspath{{figure/}}
\begin{document}

\title{Target-aware Bidirectional Fusion Transformer \\for Aerial Object Tracking}

\author{Xinglong Sun, Haijiang Sun$*$, Shan Jiang, Jiacheng Wang, Jiasong Wang 
\thanks{$*$ Corresponding author}
\thanks{Xinglong Sun, Haijiang Sun, Shan Jiang, Jiacheng Wang and Jiasong Wang are with the Changchun Institute of Optics, Fine Mechanics and Physics, Chinese Academy of Science, Changchun 130033, China (e-mail: sunxinglong@ciomp.ac.cn; sunhj@ciomp.ac.cn;jiangshan$\_$ciomp@qq.com; wangjiacheng@ciomp.ac.cn;wangjiasong@ciomp.ac.cn)}
\thanks{This work was supported by the National Natural Science Foundation of China under Grant No. 62475255.}
\thanks{Manuscript received April 19, 2021; revised August 16, 2021.}}

\markboth{IEEE GEOSCIENCE AND REMOTE SENSING LETTERS,~Vol.~xx, No.~x, August~2024}%
{Shell \MakeLowercase{\textit{et al.}}: A Sample Article Using IEEEtran.cls for IEEE Journals}


\maketitle

\begin{abstract}
The trackers based on lightweight neural networks have achieved great success in the field of aerial remote sensing, most of which aggregate multi-stage deep features to lift the tracking quality. However, existing algorithms usually only generate single-stage fusion features for state decision, which ignore that diverse kinds of features are required for identifying and locating the object, limiting the robustness and precision of tracking. In this paper, we propose a novel target-aware Bidirectional Fusion transformer (BFTrans) for UAV tracking. Specifically, we first present a two-stream fusion network based on linear self and cross attentions, which can combine the shallow and the deep features from both forward and backward directions, providing the adjusted local details for location and global semantics for recognition. Besides, a target-aware positional encoding strategy is designed for the above fusion model, which is helpful to perceive the object-related attributes during the fusion phase. Finally, the proposed method is evaluated on several popular UAV benchmarks, including UAV-123, UAV20L and UAVTrack112. Massive experimental results demonstrate that our approach can exceed other state-of-the-art trackers and run with an average speed of 30.5 FPS on embedded platform, which is appropriate for practical drone deployments.
\end{abstract}


\section{Introduction}
\IEEEPARstart{U}{AV} tracking has attracted considerable attention in visual tracking, which aims to predict the location state of a given arbitrary target in the whole UAV sequence according to its initial location. The technology gains various applications ranging from geographical survey \cite{app1}, visual localization \cite{app2} to environmental monitoring \cite{app3}. Despite great progress has been obtained, it remains challenging to achieve high-quality tracking in aerial environments due to complex interference factors, like view change, occlusion, motion blurring, etc.

With the development of deep learning, several lightweight CNNs \cite{mobilenet}, \cite{uavtrack112} are designed for aerial object tracking, which are able to realize satisfactory performance with real-time speeds. However, they have no ability to encode global and long-term semantic patterns for dealing with complex aerial scenarios. In this case, Transformer is gradually studied for UAV tracking, which is more powerful for global dependency modeling. The model has manifested great potential in executing diverse tasks, i.e., feature extraction \cite{mobilevit}, state decision \cite{a3track} and temporal modeling \cite{tctrack}. In recent years, it has been proved that aggregating multi-level features is very effective to promote the tracking quality, which is useful to provide more abundant attribute information for state prediction.

\begin{figure}[!t]
	\centering
	\includegraphics[width=0.325\linewidth]{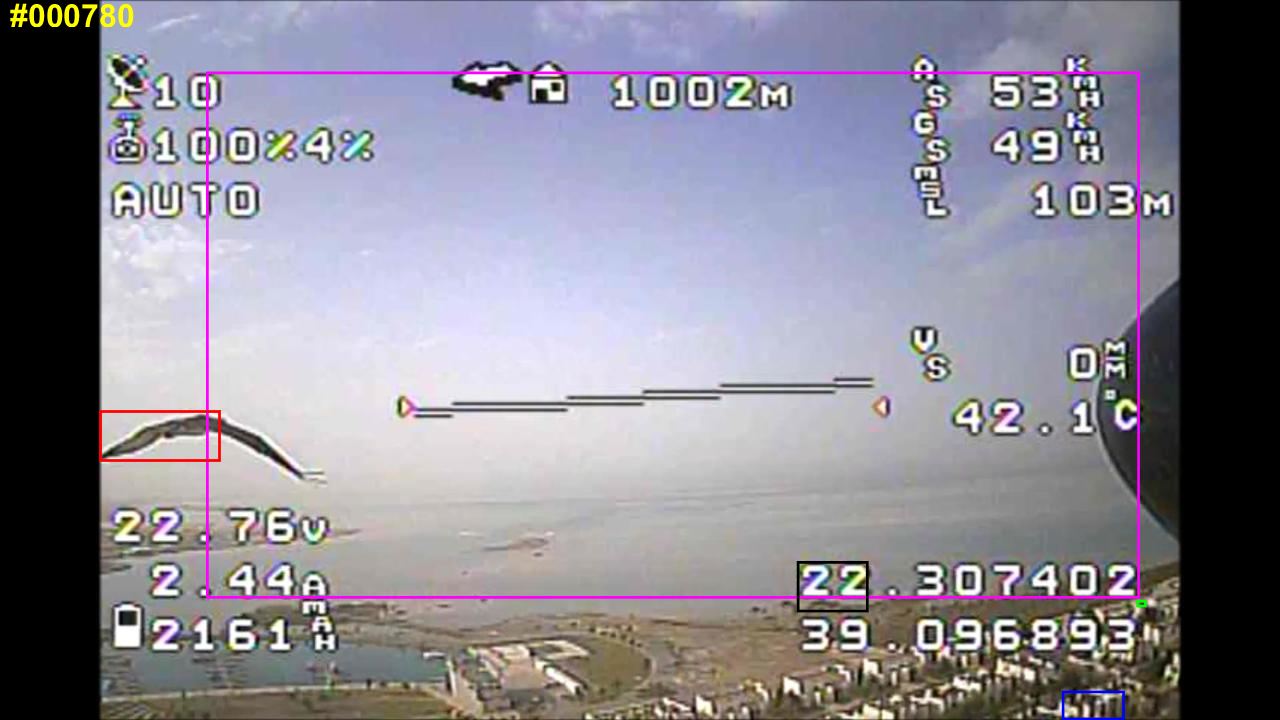}
	\includegraphics[width=0.325\linewidth]{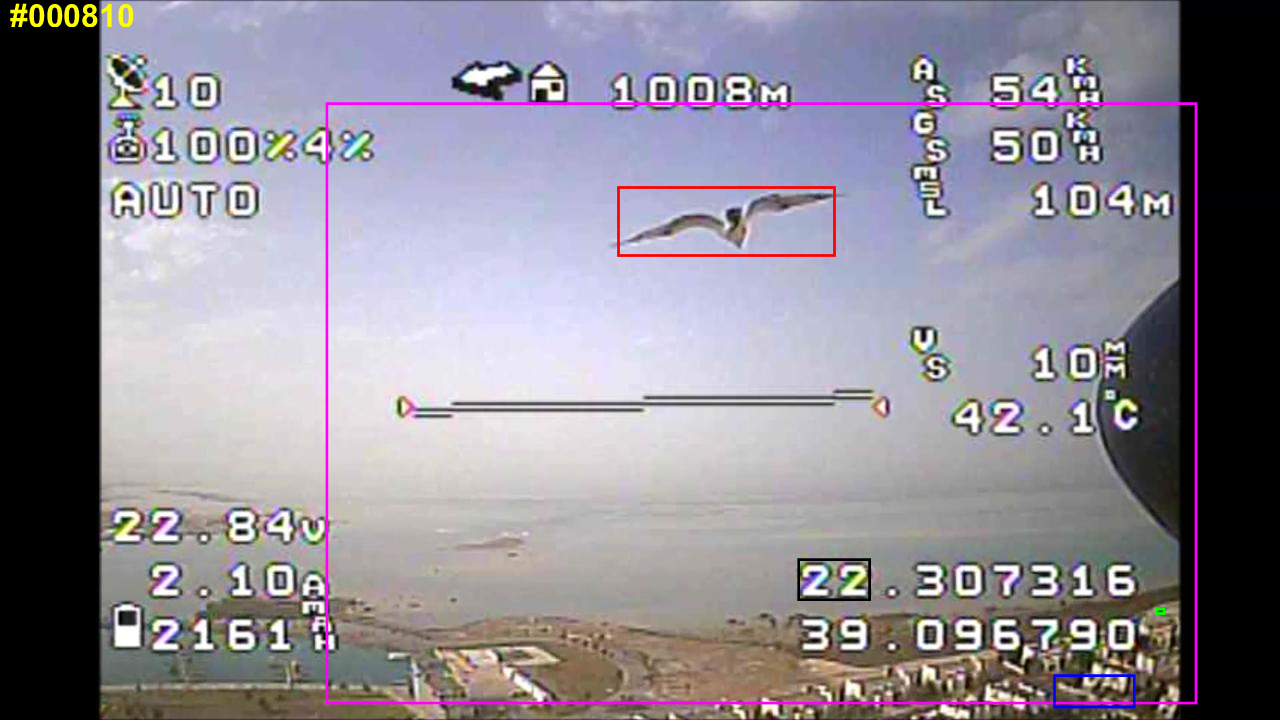}
	\includegraphics[width=0.325\linewidth]{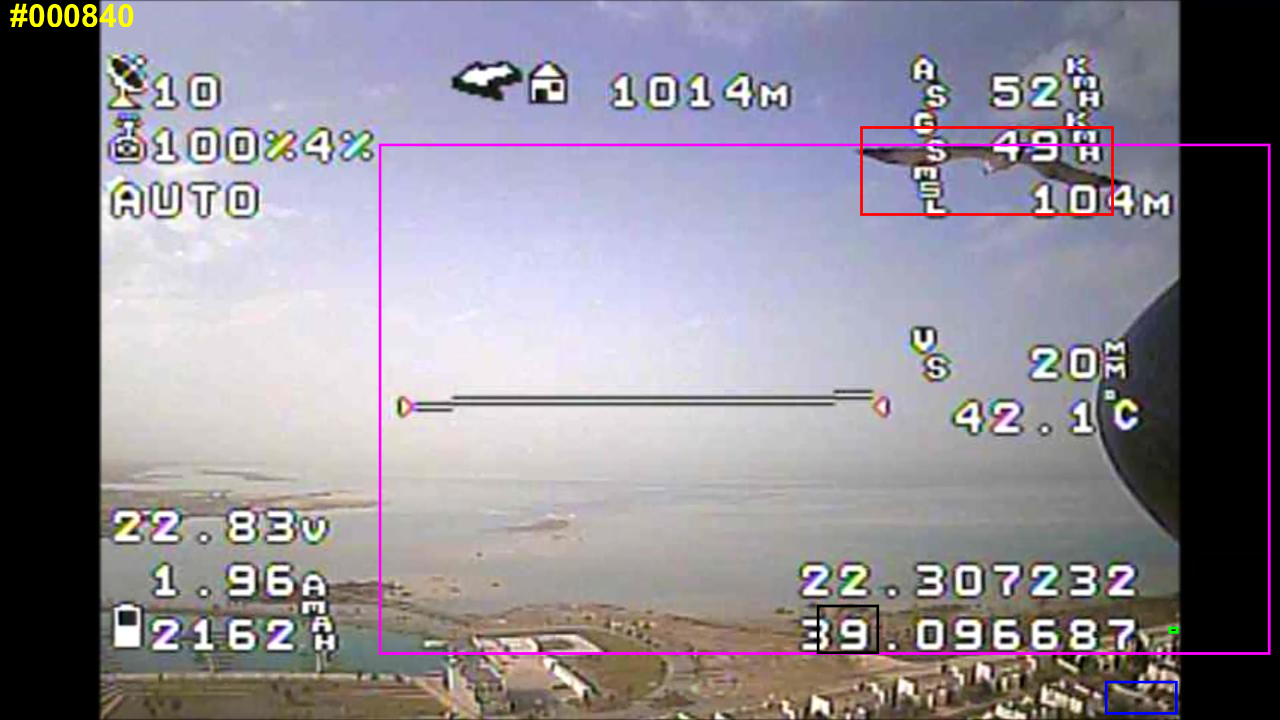}
	
	\includegraphics[width=0.325\linewidth]{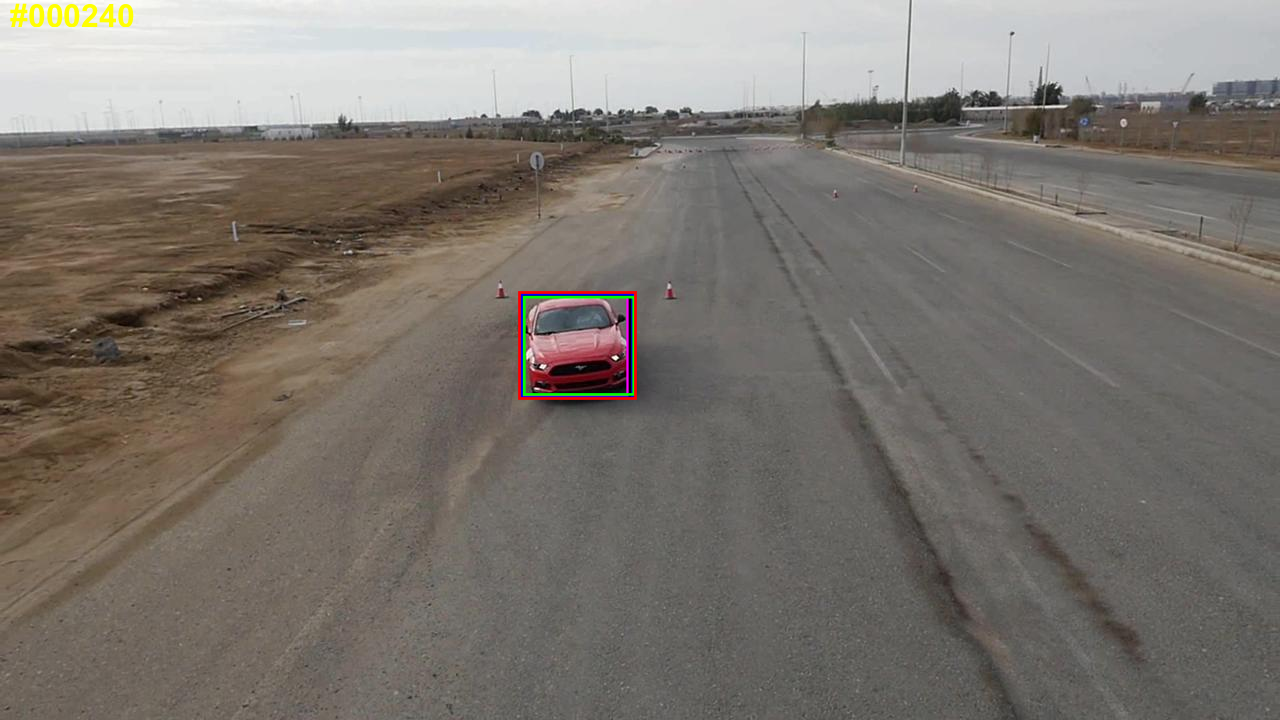}
	\includegraphics[width=0.325\linewidth]{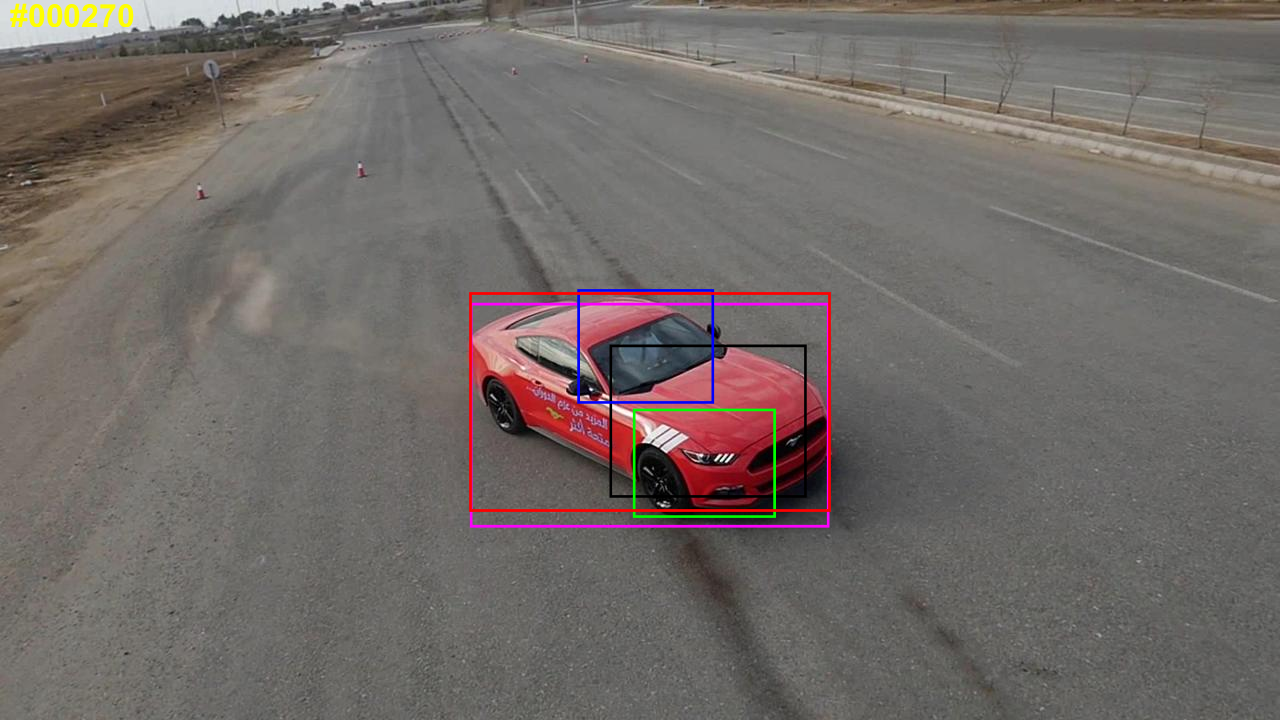}
	\includegraphics[width=0.325\linewidth]{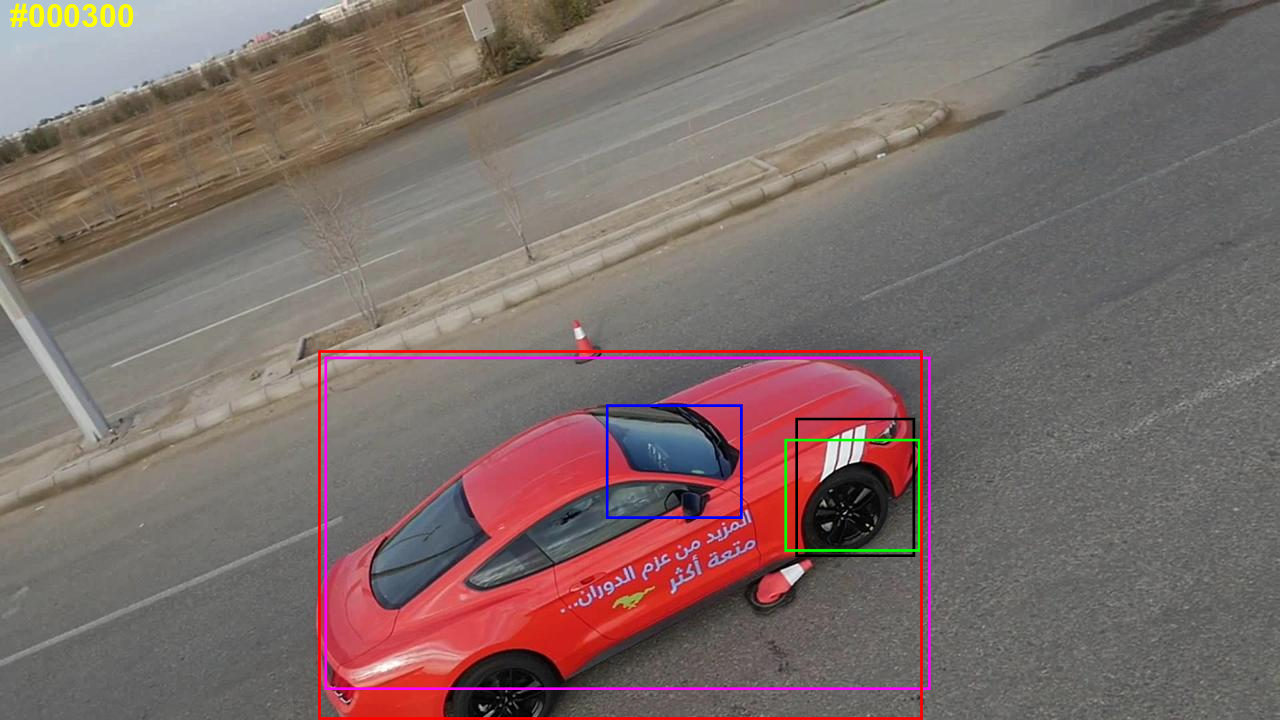}
	
	\includegraphics[width=0.325\linewidth]{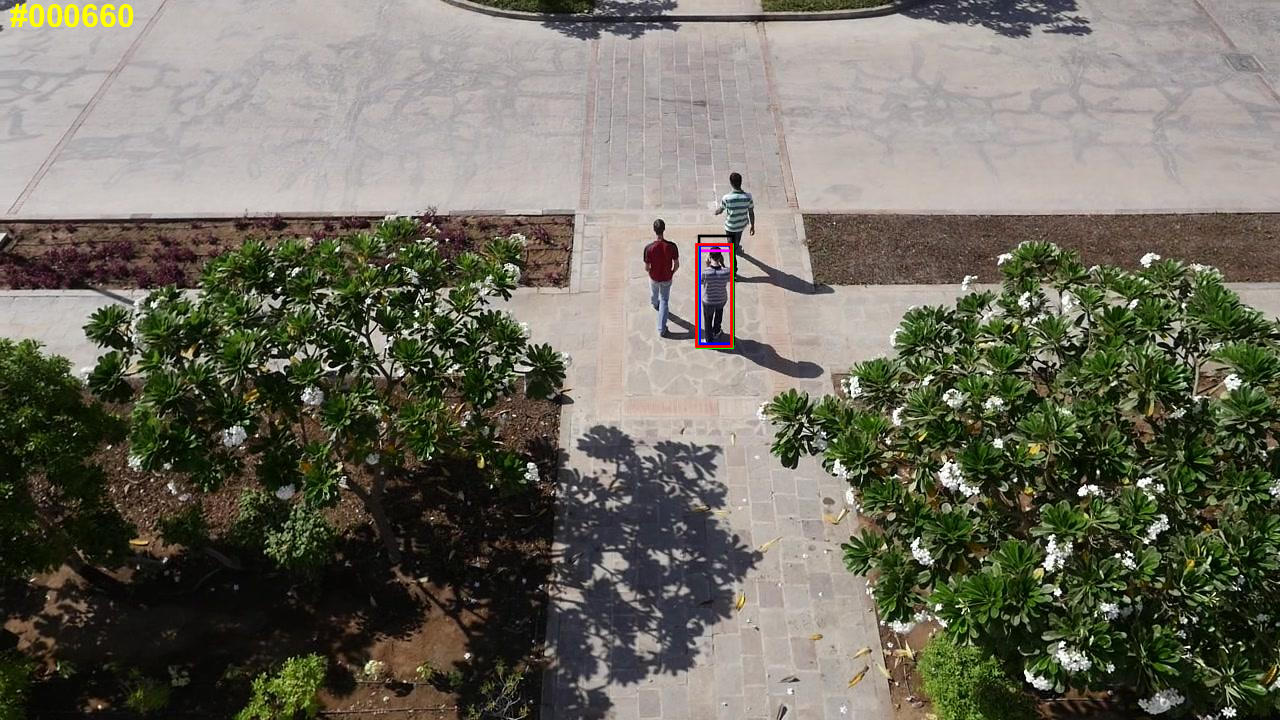}
	\includegraphics[width=0.325\linewidth]{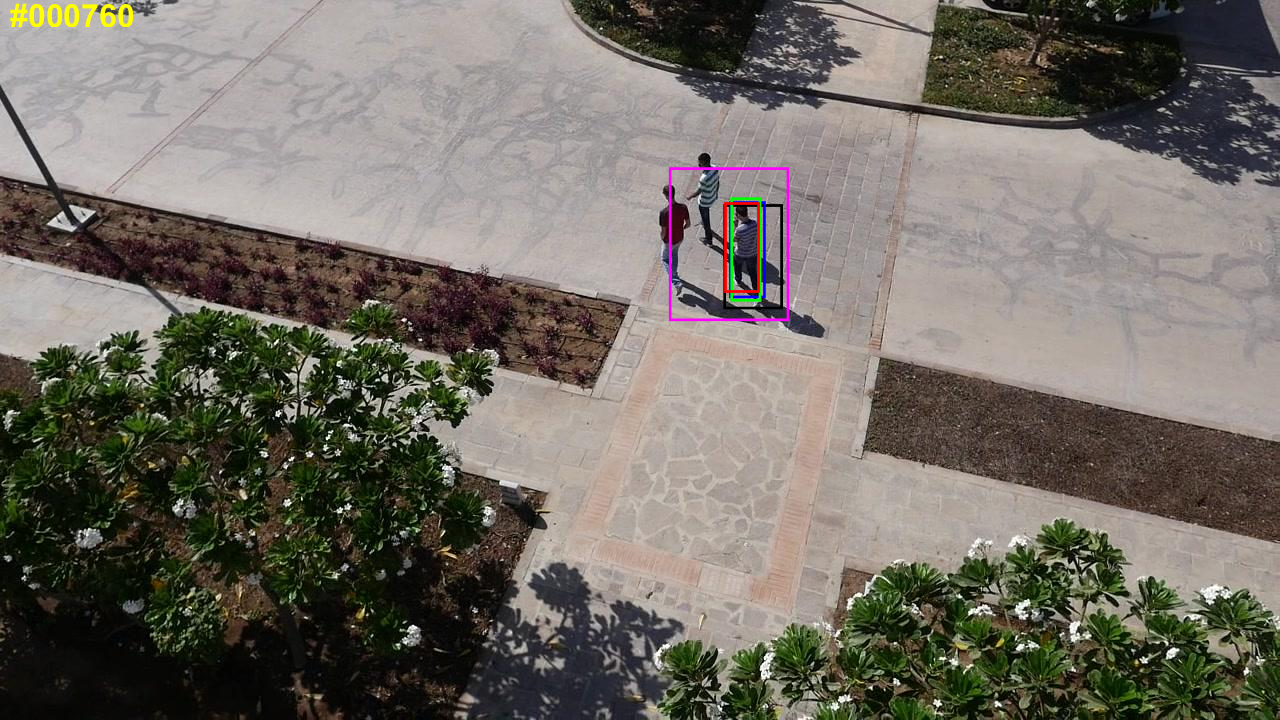}
	\includegraphics[width=0.325\linewidth]{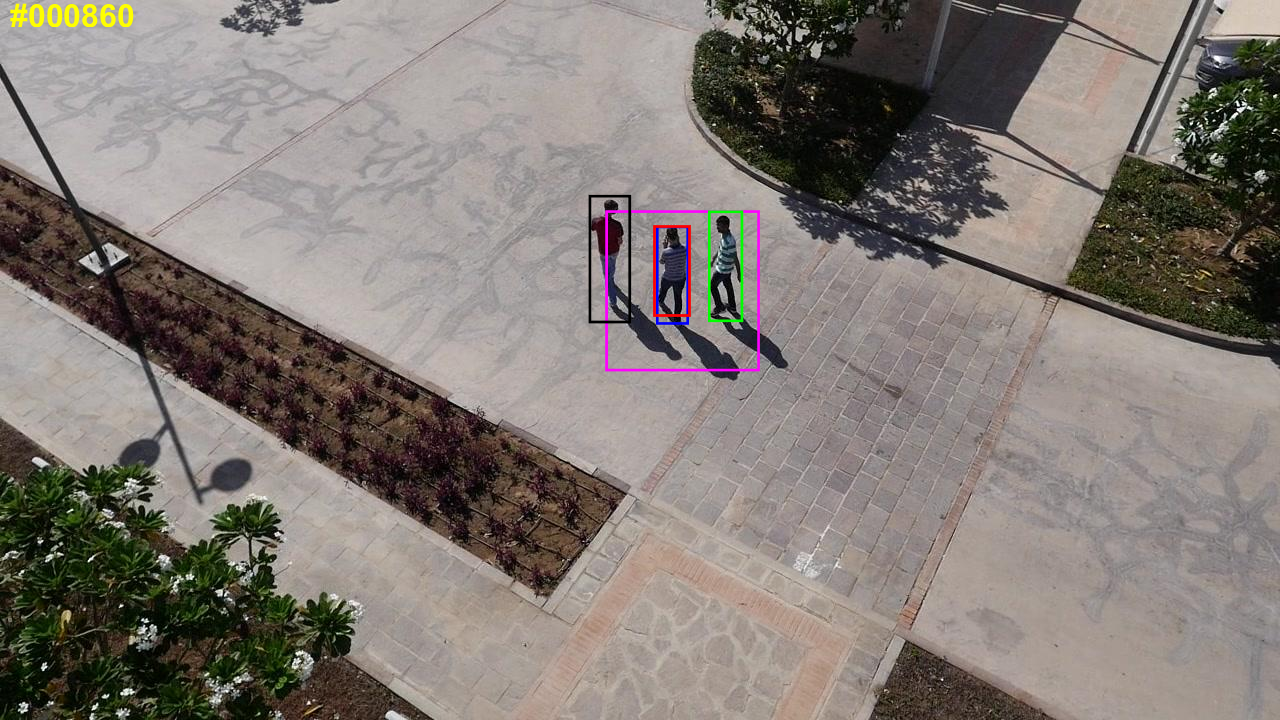}
	
	\includegraphics[width=0.95\linewidth]{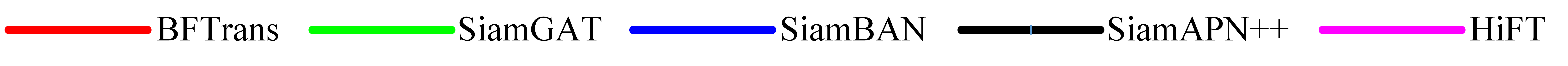}	
	\caption{Qualitative comparisons of our tracker with four state-of-the-art algorithms on several difficult sequences.}
	\label{fig_1}
\end{figure}
Various kinds of approaches have been explored to combine multi-stage expression features in a reasonable manner. HiFT \cite{hift} designed a hierarchical feature transformer to aggregate shallow and deep semantic cues for aerial tracking, following by HFPT \cite{hpft} that furtherly introduced pooling operations to improve the fusion efficiency. SiamAPN++ \cite{siamapnpp} presented a attentional aggregation network to learn the feature interdependencies adaptively, as well as SiamBRF \cite{siambrf} proposed a broad-spectrum relevance fusion network to enhance the feature representation. However, although gaining promising performance, the above works always ignore a vital issue, i.e., features with diverse attributes are simultaneously required for high-quality aerial object tracking. The shallow spatial details are valuable to lift the location precision, while the deep semantic cues are critical to discriminative the object from background. Regardless of the issue, existing fusion schemes usually only produce single-attribute aggregated features, which is inefficient to deal with complicated distractors in UAV environment.

To tackle this problem, the paper proposes a target-aware bidirectional fusion transformer for aerial tracking, which can combine the shallow and the deep features in both forward and backward process streams. The transformer would output the aggregated object features from each stream, rather than only provide single-stage fusion features for both recognition and location. Concretely, the deep features modulated by shallow details are produced to realize robust object classification, while the shallow features adjusted by deep semantics are also outputted to lift the location precision of tracker. Moreover, a target-aware positional encoding approach is presented for our fusion transformer. The method is effective in perceiving the distributions of category semantics and spatial details related to the object, fully encoding its attribute information to adapt to the appearance variations. As shown in Fig. \ref{fig_1}, the proposed BFTrans tracker can achieve better performance in some real-world UAV scenarios. The main contributions of our work are listed as follows:

1. We propose a novel target-aware bidirectional fusion transformer for aerial tracking, which allows for performing feature aggregation in a double-stream manner, providing both deep semantics for recognition and shallow details for location.

2. A target-aware positional encoding algorithm is explored to furtherly enhance the category semantics and local details of the tracked object, which is helpful to overcome complex distractors faced by UAV trackers.   

3. Massive experiments are executed on a few popular UAV benchmarks to validate the ability of the presented methods, and the results demonstrate that it exceeds other state-of-the-art approaches.

\begin{figure*}[!t]
	\centering
	\includegraphics[width=0.8\linewidth]{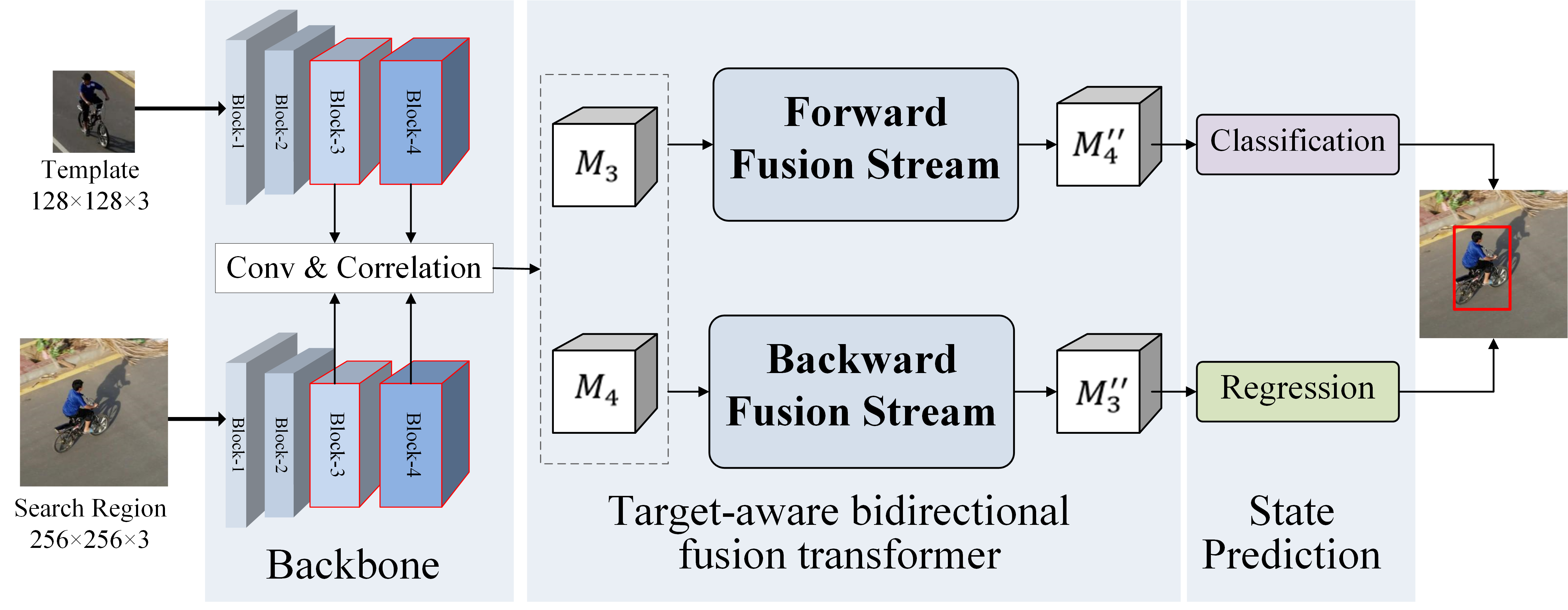}
	\caption{Overview of the proposed approach, which consists of backbone, target-aware bidirectional fusion transformer and state prediction module. The multi-stage correlation features are aggregated by both forward and backward streams of fusion transformer, generating more appropriate correlation maps for classification and regression, respectively.}
	\label{fig_2}
\end{figure*}

\section{Methodology}
\subsection{Overview}
The architecture of our presented tracker is illustrated in Fig. \ref{fig_2}. Concretely, a weight-shared backbone is first used to extract the features of template and search region, which would be compared using the correlation layers in diverse stages. Then, the multi-layer correlation features are combined by the proposed fusion transformer, encoding deep semantics and local details in two separate streams. Finally, we employ the classification and the regression heads to predict the final object state. 

\subsection{Backbone}
Following some typical cases, we use the Mobilevit-v2 \cite{mobilevit} for feature extraction, and carefully modify its structure to lift the efficiency and adaptability on real UAV platforms. Firstly, the last fifth block is removed to reduce the computing complexity. Moreover, we append the $3\times3$ convolution with stride 2 and $1\times1$ convolution with stride 1 to the third and the fourth blocks respectively, which would be used to output the multi-stage features with identical size. For the features of template $\phi_k(z) \in \mathbb{R}^{H_z \times W_z \times C_k}$ and candidate region $\phi_k(x) \in \mathbb{R}^{H_x \times W_x \times C_k}$ from $k$-th stages, the pixel-wise cross-correlation is employed to compute the similarity responses $M_k$:
\begin{equation} \label{deqn_ex1a}
M_k=\mathrm{C}_k\left(\phi_k(z) \odot \phi_k(x)\right), \quad k=3,4
\end{equation}
in which $\odot$ denotes the pixel-wise cross-correlation operation and $C_k$ is a $1\times1$ convolutional layer to normalize the channel to $d$ ($d=192$). 

\begin{figure}[!t]
	\centering
	\includegraphics[width=0.95\linewidth]{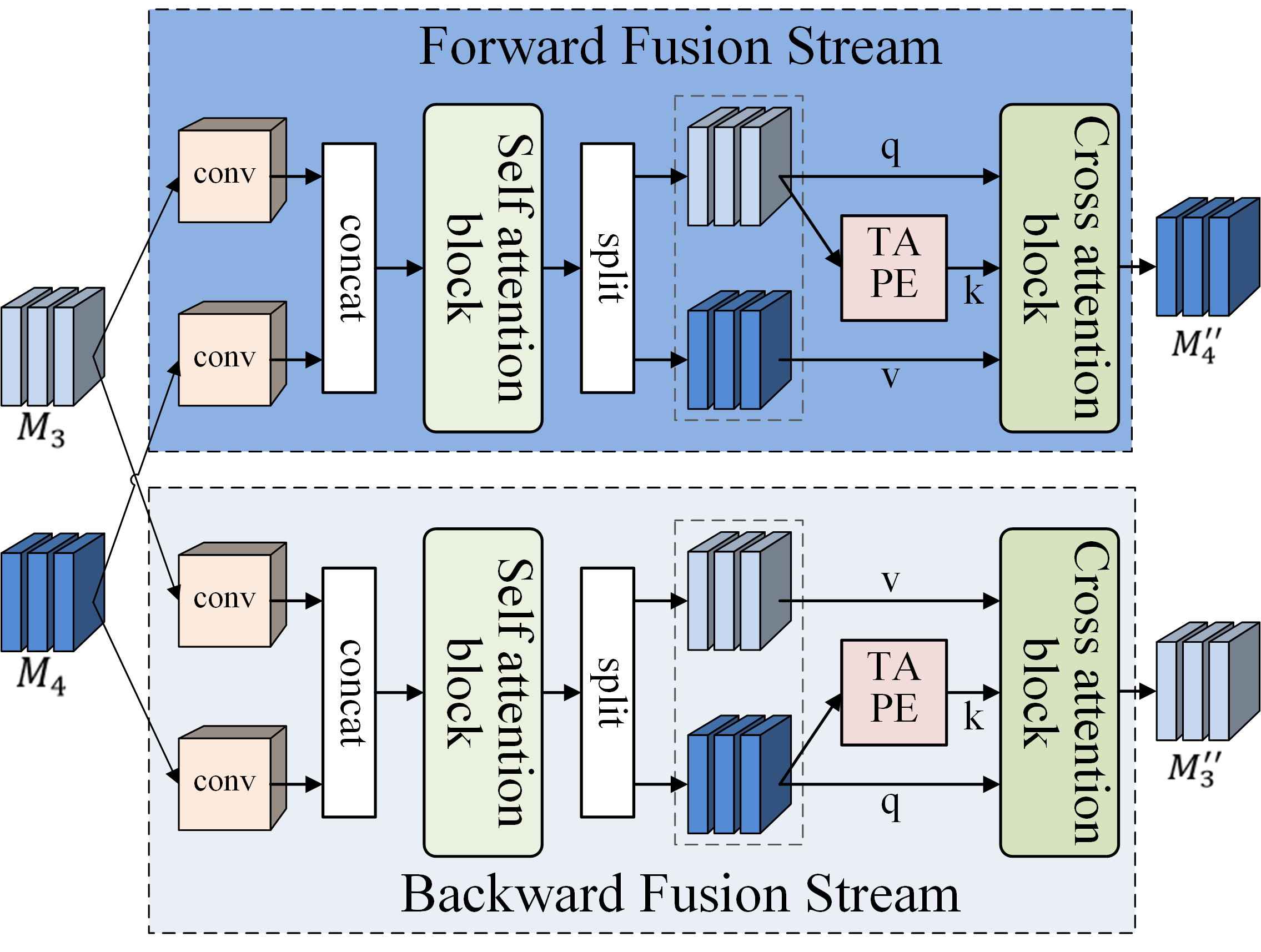}
	\caption{Structure of our target-aware bidirectional fusion transformer, in which both self and cross attention blocks are utilized to analyze features.}
	\label{fig_3}
\end{figure}

\subsection{Feature Fusion Transformer}
Considering that previous fusion models are not sufficient to ensure the tracking quality, we present the bidirectional-fusion transformer and the target-aware positional encoding strategy in this part, as expounded as follows. 

\subsubsection{Bidirectional Fusion Transformer}
The key of feature fusion is to provide detailed cues and semantic patterns for state decision simultaneously. To realize the goal, we deploy both the forward and the backward fusion streams in the proposed transformer, whose network structure is depicted in Fig. \ref{fig_3}. In each stream, two $1\times1$ convolutional layers are first adopted to fine-tune the shallow and the deep correlation features, which are concatenated to get the joint features $M_c$ for self-attention blocks. The block takes a linear self-attention \cite{mobilevit} to model the dependencies between two kinds of features, following by a feed-forward network to fully encode the attribute relationships. The role of a self-attention block can be formulated as:
\begin{equation} \label{deqn_ex2}
M_c^{\prime}=M_c+\mathcal{F}\left(A_{\text {self }}\left(Q_c, K_c, V_c\right)\right)
\end{equation}
where, $A_{\text {self }}$ and $\mathcal{F}$ denote the linear self-attention and the feed-forward network, respectively. After splitting $M_c^{\prime}$ into shallow features $M_3^{\prime}$ and deep features  $M_4^{\prime}$, which are furtherly aggregated by cross-attention blocks. In the forward stream, the linear cross-attention employs $M_3^{\prime}$ as the query and the key tokens to adjust the value token of  $M_4^{\prime}$, which is also followed by a feed-forward module to output the final fused results. On the contrary, the deep correlation features are utilized to adjust shallow features in the backward fusion stream, which is very critical to promote the regression accuracy. In the above steps, the object-aware positional encoding (TAPE) is applied to key tokens for better perceiving appearance attributes, which will be described in the followed section. The functions of cross-attention block can be formulated as:
\begin{equation} \label{deqn_ex3}
M_4^{\prime \prime}=M_4^{\prime}+\mathcal{F}\left(A_{\text {cross }}\left(Q_3^{\prime}, K_3^{\prime}, V_4^{\prime}\right)\right)
\end{equation}
\begin{equation} \label{deqn_ex4}
M_3^{\prime \prime}=M_3^{\prime}+\mathcal{F}\left(A_{\text {cross }}\left(Q_4^{\prime}, K_4^{\prime}, V_3^{\prime}\right)\right)
\end{equation}
where, $A_{\text {cross}}$ denotes the linear cross-attention unit. With our fusion network, sufficient semantic cues and detailed patterns can be obtained for recognition and regression, respectively. 

\subsubsection{Target-aware Positional Encoding}  \label{sect_tape}
Self or cross attention is permutation-invariant, since some positional encoding strategies are often adopted to model the relative positional relationships. However, classical encoding methods, like sinusoidal positional encoding, are completely unrelated with the inputted features, which lack the ability to learn the object appearance attributes. In addition, they are generally scale-sensitive, which are not suitable for aerial tracking task where the scale of interested object may vary dramatically.

\begin{figure}[!t]
	\centering
	\includegraphics[width=0.9\linewidth]{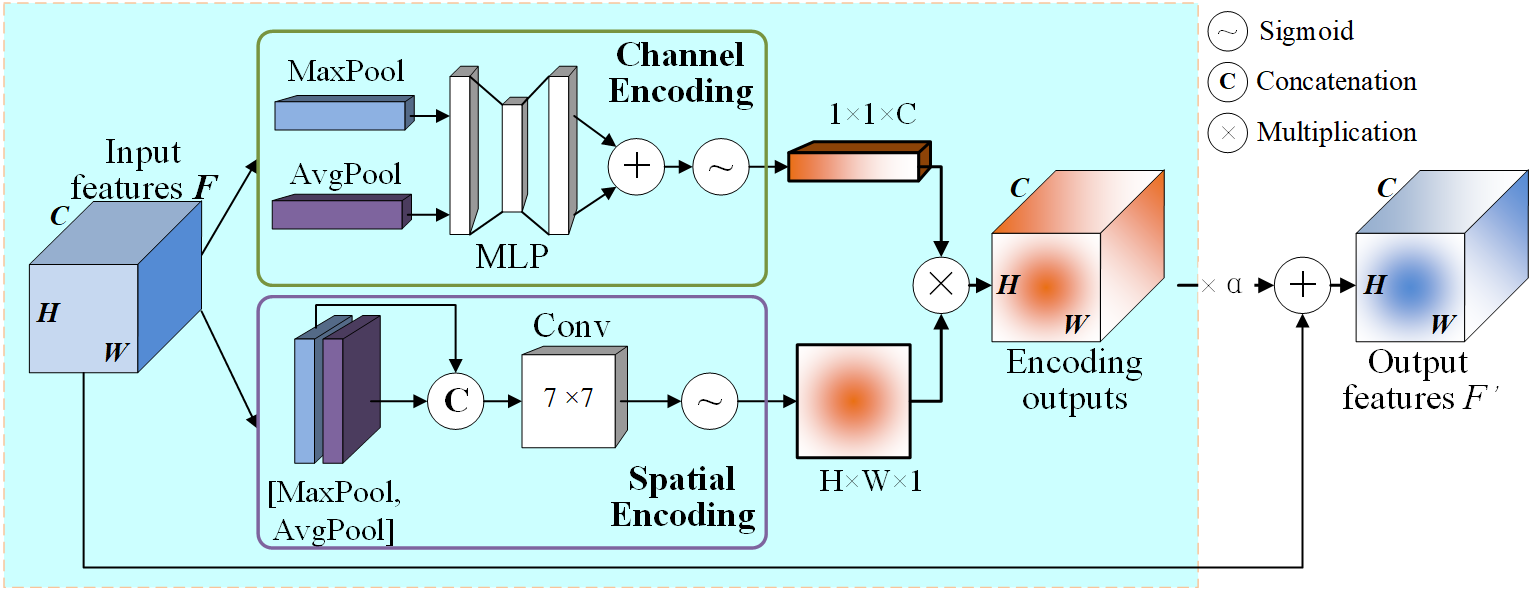}
	\caption{Framework of target-aware positional encoding module, mainly containing the channel and the spatial encoding blocks.}
	\label{fig_4}
\end{figure}

To address the above drawbacks, we present a target-aware positional encoding scheme, which is mainly comprised by the channel and the spatial encoding blocks, as illustrated in Fig. \ref{fig_4}. Channel encoding block \cite{cbam} first uses average and maxing global pooling layers to compress the feature sizes, following by a multi-layer perception ($MLP$) to analyze the pooling features. Then, two kinds of pooling features are accumulated, and the result is normalized by a sigmoid function. In the spatial encoding block \cite{cbam}, after reducing the channel quantity of features with average and max channel pooling layers, a convolutional layer is imposed on the concatenated pooling features to capture the local spatial attributes, where the sigmoid layer is also used to normalize the encoding result. At last, the results of channel and spatial encoding are multiplied to obtain the final positional encoding responses, which will be adopted to update the original features in a weight average manner. The proposed positional encoding algorithm can be formulated as:

\begin{equation} \label{deqn_ex5}
W_c=g\left(M L P\left(\operatorname{Pool}_{\max }^{h w}(F)+\operatorname{Pool}_{\mathrm{avg}}^{h w}(F)\right)\right) 
\end{equation}
\begin{equation} \label{deqn_ex6}
W_s=g\left(\operatorname{Conv}\left(\operatorname{Pool}_{\max }^c(F) ; \operatorname{Pool}_{\mathrm{avg}}^c(F)\right)\right) 
\end{equation}
\begin{equation} \label{deqn_ex7}
F^{\prime}=F+\alpha \cdot\left(W_c \odot W_s\right)
\end{equation}
in which, $\operatorname{Pool}_{\mathrm{max}}^{h w}$ and $\operatorname{Pool}_{\mathrm{avg}}^{h w}$ represent the max and the average spatial pooling layers, respectively, while $\operatorname{Pool}_{\mathrm{max}}^{c}$ and $\operatorname{Pool}_{\mathrm{avg}}^{c}$ denote the max and the average pooling operations along channel dimension. $g$ is the sigmoid layer, and $\alpha$ is a learnable weight factor. By introducing our encoding scheme, the fusion model can learn the object attribute information from diverse aspects, lifting the adaptability of online tracking.

\subsection{State Prediction}
State prediction is performed by one classification head and two regression heads, all of which are composed of a few stacked convolutional layers. Given the fused deep correlation features $M_4^{\prime \prime}$, the classification head computes the foreground score of each candidate element. At the same time, two regression heads estimate the normalized center offsets and object sizes of all elements based on the shallow correlation features $M_3^{\prime \prime}$, respectively. For optimization, we employ the weighted focal function as classification loss, and accumulate the $l_1$-norm loss and the generalized IoU loss to train the regression branch. The overall loss function is formulated as:
\begin{equation} \label{deqn_ex8}
\mathcal{L}_T=\mathcal{L}_{f o c a l}+\lambda_1 \cdot \mathcal{L}_{l 1}+\lambda_2 \cdot \mathcal{L}_{G I o U}
\end{equation}
in which, $\mathcal{L}_{f o c a l}$, $\mathcal{L}_{l 1}$ and $\mathcal{L}_{G I o U}$ depicts the focal loss, $l_1$-norm loss and the generalized IoU loss. $\lambda_1$ and $\lambda_2$ are the weight factors for balancing diverse kinds of loss functions, which are set to 2 and 5, respectively.

\section{Experiments and Results}
\subsection{Implementation Details}
The presented tracking model is optimized on the data splits of LaSOT \cite{lasot}, GOT-10k \cite{got10k} TrackingNet \cite{trackingnet} and COCO \cite{coco}, in which the sizes of template and search region are set to $128 \times 128$ and $256 \times 256$, respectively. During optimization, the backbone i.e., Mobilevit-v2, is first initialized with the weights provided by \cite{mobilevit}. We adopt a AdamW optimizer with the weight decay of 1e-4 to train the proposed network for 300 epochs, where the batch size is equal to 128 and every epoch consists of 60000 sample pairs. The initial learning rates of backbone are set to 4e-5, while the learning rates of other modules are 4e-4, all of which decrease 10 times after 240 epochs. The proposed method is executed on one NVIDIA RTX 4090 GPU with Pytorch 1.13, and the whole training phase takes about 13 hours. 

\subsection{Benchmarks and Metrics}
The proposed tracking approach is evaluated on three public UAV benchmarks: UAV123 \cite{uav123}, UAV20L \cite{uav123} and UAV-Track112 \cite{uavtrack112}. UAV123 is a typical aerial tracking dataset including 123 sequences, which cover different challenging scenarios, i.e., illumination variation, occlusion, background clutter, etc. UAV20L contains 20 long-term video sequences, which is a great challenge to short-term UAV trackers. There
are 112 sequences in the UAVTrack112 dataset, all of which are captured from low-attitude unmanned aerial vehicles.

In the evaluation protocols of these datasets, Success Rate and Precision Rate are used to evaluate trackers. Success Rate denotes the Area Under Curve (AUC) of success plot which is the percentage of images when the overlap ratios are larger than a given threshold. Precision Rate is the ratio of images when the distance errors are within a given threshold, which is usually set to 20 pixels.
 
\begin{table*}[t]
	\caption{Comparisons with several state-of-the-art trackers on three public and popular benchmarks, in which both performance and speed are considered.  The best three results are highlighted in \textcolor[rgb]{1,0,0}{red}, \textcolor[rgb]{0,1,0}{green} and \textcolor[rgb]{0,0,1}{blue} fonts.}	\label{first-tab}	
	\begin{center}
		\setlength{\tabcolsep}{2.5mm}{		
			\begin{tabular}{lc|ccccccccc}
				\toprule[1.5pt] 
				Datasets& &HiFT & SiamAPN++ & SGDViT & PRL-Track & LightTrack & Ocean & SiamBAN & SiamGAT & Ours\\
				\midrule[1pt] 
				\multirow{2}*{UAV123} & 
				    Succ. & 0.589 & 0.582 & 0.581 & 0.593 & --    & 0.621 & \textcolor[rgb]{0,0,1}{0.631} &\textcolor[rgb]{0,1,0}{0.640} &\textcolor[rgb]{1,0,0}{0.647} \\
				~ & Prec. & 0.787 & 0.768 & 0.763 & 0.791 & --    & 0.823 & \textcolor[rgb]{0,0,1}{0.833} & \textcolor[rgb]{0,1,0}{0.835} &\textcolor[rgb]{1,0,0}{0.847} \\
				\midrule[1pt] 
				\multirow{2}*{UAV20L} &
				 	Succ. & 0.553 & 0.533 & 0.519 & --    & \textcolor[rgb]{0,1,0}{0.620} & 0.444 & 0.564 & \textcolor[rgb]{0,1,0}{0.620} & \textcolor[rgb]{1,0,0}{0.638} \\
				~ & Prec. & 0.736 & 0.703 & 0.692 & --    & \textcolor[rgb]{0,0,1}{0.791} & 0.630 & 0.736 & \textcolor[rgb]{0,1,0}{0.796} &\textcolor[rgb]{1,0,0}{0.825} \\
				\midrule[1pt] 
				\multirow{2}*{UAVTrack112} & 
					Succ. & 0.570 & 0.586 & 0.599 & 0.602 & 0.619 & 0.518 & \textcolor[rgb]{0,1,0}{0.625} & \textcolor[rgb]{0,0,1}{0.620} &\textcolor[rgb]{1,0,0}{0.650} \\
				~ & Prec. & 0.742 & 0.770 & 0.770 & 0.786 & 0.783 & 0.713 & \textcolor[rgb]{0,1,0}{0.813} & \textcolor[rgb]{0,0,1}{0.799} &\textcolor[rgb]{1,0,0}{0.821} \\
				\midrule[1pt] 
				Speed (Xavier) & &\textbf{31.2} & \textbf{35.7} & 23.2 & \textbf{35.6} & 21.2 & 16.8 & 6.3 & 17.4 & \textbf{30.5}\\
				\bottomrule[1.5pt]
		\end{tabular}}
	\end{center}
\end{table*}

\subsection{Comparison with State-of-the-Art Trackers}
To present the superiority of the proposed method, we compare it with eight outstanding trackers, consisting of HiFT \cite{hift}, SiamAPN++ \cite{siamapnpp}, SGDViT \cite{sgdvit}, PRL-Track \cite{prltrack}, LightTrack \cite{lighttrack}, Ocean \cite{ocean}, SiamBAN \cite{siamban} and SiamGAT \cite{siamgat}. The tracking results are reported on Table \ref{first-tab}, in which the tracking speed is tested on NVDIA Jetson AGX Xavier.

\subsubsection{UAV123}
On the benchmark, the proposed tracker achieves the best performance with the highest Success and Precision scores of 64.7\% and 84.7\%. Compared to the typical SiamBAN tracker, our BFTrans outperforms it by 1.6\% on Success and 1.4\% on Precision, as well as is about 6 times faster than the tracker. In addition, our method is superior to another aerial tracker, i.e., PRL-Track, by 5.4\% on Success and 5.6\% on Precision. These results manifests that the proposed fusion scheme is effective to lift the tracking ability in complex aerial scenarios. 
\subsubsection{UAV20L}
The presented method realizes very excellent performance on both evaluation metrics. In comparison with the second-ranked SiamGAT, our tracker produces great gains of 1.8\% on Success and 2.9\% on Precision. In the other aspect, our algorithm exhibits a remarkable advantage in terms of speed. In addition, our tracking model surpasses the state-of-the-art UAV tracker of SiamAPN++ by 10.5\% on Success and 12.3\% on Precision.

\subsubsection{UAVTrack112}
Our BFTrans tracker obtains very outstanding performance and performs favorably against all of comparison trackers on both Success and Precision scores. Comparison to the light-weight tracking model of LightTrack, our works yields great increments on two metrics with higher computing efficiency, declaring the great advantages of the proposed network model.

\begin{figure}[!t]
	\centering
	\includegraphics[width=0.8\linewidth]{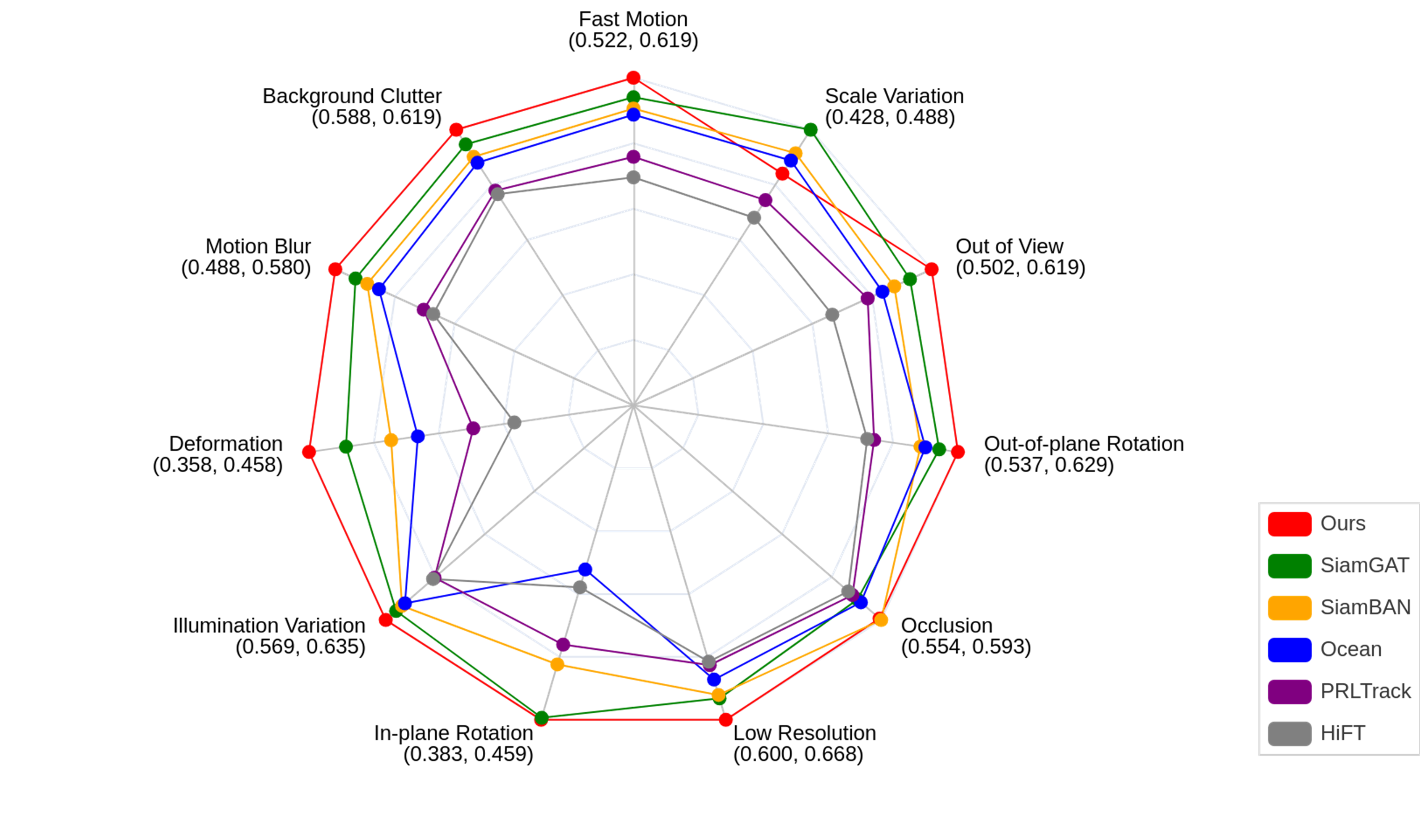}
	\caption{Success scores of diverse attributes on UAV123 dataset, where the values in parentheses depicts the minimum and maximum scores of all trackers on each attribute.}
	\label{fig_5}
\end{figure}

\subsection{Attribute-based Analysis}
To furtherly demonstrate the concrete performance of our method, we conduct the attribute-based comparison on UAV123 dataset, in which BFTrans is evaluated on 11 challenging attributes. As shown in Fig. \ref{fig_5}, the presented tracker is able to realize remarkable performance on all complex scenarios, and ranks first on 9 different attributes in terms of Success. Specifically, our work outperforms the second-ranked method with more than 1.0\% increments on 8 attributes, including Fast motion (FM), Background clutter (BC), Deformation (DEF) and so on. These phenomena prove that the proposed fusion network is useful to learn more appropriate features for instance classification and location regression. 

\begin{table}[h]
	\caption{Ablation studies of diverse variations on UAV123 datasets.The best results are highlighted in \textcolor[rgb]{1,0,0}{red} fonts.}	\label{second-tab}	
	\begin{center}
		\setlength{\tabcolsep}{4.5mm}{		
			\begin{tabular}{l|cc|c|cc}
				\toprule[1.5pt] 
				$\#$  & FFM & BFM & TAPE & Succ. & Prec.\\
				\midrule[1pt] 
				1  &  &  &  & 0.596 & 0.799\\
				2  & \checkmark &  &  & 0.613 & 0.812\\
				3  &  & \checkmark &  & 0.610 & 0.807\\
				4  & \checkmark & \checkmark &  & 0.634 & 0.831\\
				5  & \checkmark & \checkmark & \checkmark & \textcolor[rgb]{1,0,0}{0.647} & \textcolor[rgb]{1,0,0}{0.847}\\
				\bottomrule[1.5pt]
		\end{tabular}}
	\end{center}
\end{table} 

\subsection{Ablation studies} 
We conduct massive ablation experiments by setting up five ablation variations to verify the impact of each contribution. Concretely, we first construct a baseline tracker in Variant $\#$1, which adopts our backbone and directly estimates the object state with the correlation maps from the fourth network block. In Variations $\#$2 and $\#$3, the proposed Forward Fusion Module (FFM) and Backward Fusion Module (BFM) are introduced alternately. In contrast to them, Variation $\#$4 introduces both FFM and BFM to form the bidirectional-fusion scheme. Variant $\#$5 furtherly incorporates the presented target-aware positional encodings (TAPE), formatting our BFTrans tracker.

As reported on Table \ref{second-tab}, it is clearly seen that both FFM and BFM can improve the tracking ability of the baseline obviously. Compared to Variant $\#$2, Variant $\#$4 is superior to it by 2.1\% on Success and 1.9\% on Precision, which declares that the proposed bidirectional fusion transformer is more effective for robust classification and precise regression. Moreover, Variant $\#$5 surpasses Variant $\#$4 by 1.3\% on Success and 1.6\% on Precision, declaring that our TAPE strategy is important to lift the tracking performance on UAV environments.

\section{CONCLUSION}
In this work, we proposed a novel target-aware bidirectional fusion transformer for aerial tracking. Firstly, both the forward and the backward fusion streams are embedded into a unified transformer model, which is very valuable for capturing the local details for precise location and the global semantics for robust classification simultaneously. Moreover, we designed a target-aware positional encoding scheme for the fusion transformer, which can encode more object appearance attributes to ensure tracking stability. Massive experimental results on three recent datasets manifested that the proposed BFTrans tracker is able to achieve the leading performance with the real-time tracking speed. 


\end{document}